\documentclass[letterpaper, 10 pt, journal, twoside]{IEEEtran} 
\usepackage{amsmath,amsfonts}

\DeclareMathOperator*{\argmin}{arg\,min}
\usepackage{algorithmic}
\usepackage{algorithm}
\usepackage{array}
\usepackage[caption=false,font=normalsize,labelfont=sf,textfont=sf]{subfig}
\usepackage{stfloats}
\usepackage{url}
\usepackage{verbatim}
\usepackage{graphicx}
\usepackage{cite}
\usepackage{newtxtext,newtxmath}
\usepackage{caption}
\usepackage{hyperref}
\usepackage{color}
\usepackage{todonotes}
\usepackage{footnote}
\newcommand{\revise}[1]{\textcolor{black}{#1}} 

\begin{document}
	
	\title{\LARGE \bf Deep Koopman Operator with Control for Nonlinear Systems}
	
	\author{Haojie Shi, \textit{Graduate Student Member, IEEE},  Max Q.-H. Meng$^{\dag}$, \textit{Fellow, IEEE}
		\thanks{Manuscript received: February 13, 2022; Revised May 14, 2022; Accepted May 29, 2022.}
\thanks{This paper was recommended for publication by Editor Dana Kulic upon evaluation of the Associate Editor and Reviewers’ comments. This work was partially supported by National Key R\&D program of China
	with Grant No. 2019YFB1312400, Hong Kong RGC CRF grant C4063-18G, Hong Kong RGC GRF grant No. 14211420 and Hong Kong RGC TRS grant T42-409/18-R awarded to Max Q.-H. Meng.  \textit{(Corresponding author: Max Q.-H. Meng)}}
\thanks{Haojie Shi is from the Chinese University of Hong Kong, Hong Kong,
	{\tt\footnotesize h.shi@link.cuhk.edu.hk}.}
\thanks{Max Q.-H. Meng is with Shenzhen Key Laboratory of Robotics Perception and Intelligence, and the Department of Electronic and Electrical Engineering, Southern University of Science and Technology, Shenzhen 518055, China, on leave from the Department of Electronic Engineering, The Chinese University of Hong Kong, Hong Kong, and also with the Shenzhen Research Institute of The Chinese University of Hong Kong, Shenzhen 518057, China,  {\tt\footnotesize max.meng@ieee.org}}
\thanks{Digital Object Identifier (DOI): see top of this page.}}
	
		\markboth{IEEE Robotics and Automation Letters. Preprint Version. Accepted MAY, 2022}
	{SHI \MakeLowercase{\textit{et al.}}: DEEP KOOPMAN WITH CONTROL} 
	
	
	\maketitle
	
	\begin{abstract}
		Recently Koopman operator has become a promising data-driven tool to facilitate real-time control for unknown nonlinear systems. It maps nonlinear systems into equivalent linear systems in embedding space, ready for real-time linear control methods. However, designing an appropriate Koopman embedding function remains a challenging task. Furthermore, most Koopman-based algorithms only consider nonlinear systems with linear control input, resulting in lousy prediction and control performance when the system is fully nonlinear with the control input. In this work, we propose an end-to-end deep learning framework to learn the Koopman embedding function and Koopman Operator together to alleviate such difficulties. We first parameterize the embedding function and Koopman Operator with the neural network and train them end-to-end with the K-steps loss function. \revise{ Then, an auxiliary control network is augmented to encode the nonlinear state-dependent control term to model the nonlinearity in the control input. This encoded term is considered the new control variable instead to ensure linearity of the modeled system in the embedding system.} \revise{
	We next} deploy Linear Quadratic Regulator (LQR) on the linear embedding space to derive the optimal control policy and decode the actual control input from the control net. Experimental results demonstrate that our approach outperforms other existing methods, reducing the prediction error by \revise{order of magnitude} and achieving superior control performance in several nonlinear dynamic systems like damping pendulum, CartPole, and  \revise{ the seven DOF} robotic manipulator. 
\end{abstract}


	\section{INTRODUCTION}
\IEEEPARstart{T}{he} prediction and control of nonlinear systems remain a challenging task. Classical methods first require full knowledge of the system model to predict the evolution of dynamical systems and then design state feedback control laws\cite{sastry2013nonlinear} or optimization-based control policies\cite{li2004iterative,allgower2012nonlinear}. However, the hidden dynamics of chaotic nonlinear systems are usually unknown or too complicated to be modeled, prohibiting such controllers from taking effect.

For the prediction of nonlinear systems, great research interest has been put into data-driven methods like SINDy \cite{brunton2016sparse}. The deep neural network also plays an important role in nonlinear system modeling with its strong expressive power\cite{kumar2019comparative}. Both of these methods discover hidden dynamics in a nonlinear form, demanding nonlinear control methods such as iLQR\cite{li2004iterative} and NMPC\cite{allgower2012nonlinear}. \revise{Nevertheless,} these optimization-based methods require tremendous computational time with the increasing dimensions of the system state, \revise{which is} prohibitive for real-time control of high dimensional nonlinear systems. Besides, model-based RL approaches illustrate great success in controlling nonlinear systems. They first train the neural network to approximate the dynamics and then apply policy search methods for trajectory optimization via backpropagation\cite{wang2019benchmarking,tassa2012synthesis}. Model-free RL approaches also achieve strong performance on continuous control of nonlinear systems\cite{silver2014deterministic,silver2014deterministic,haarnoja2018soft}. However, these methods suffer from sample efficiency issues and lack of generalization.
\begin{figure}[t!]
	\vspace{-.3cm}
	\centering
	\includegraphics[width=0.5\textwidth]{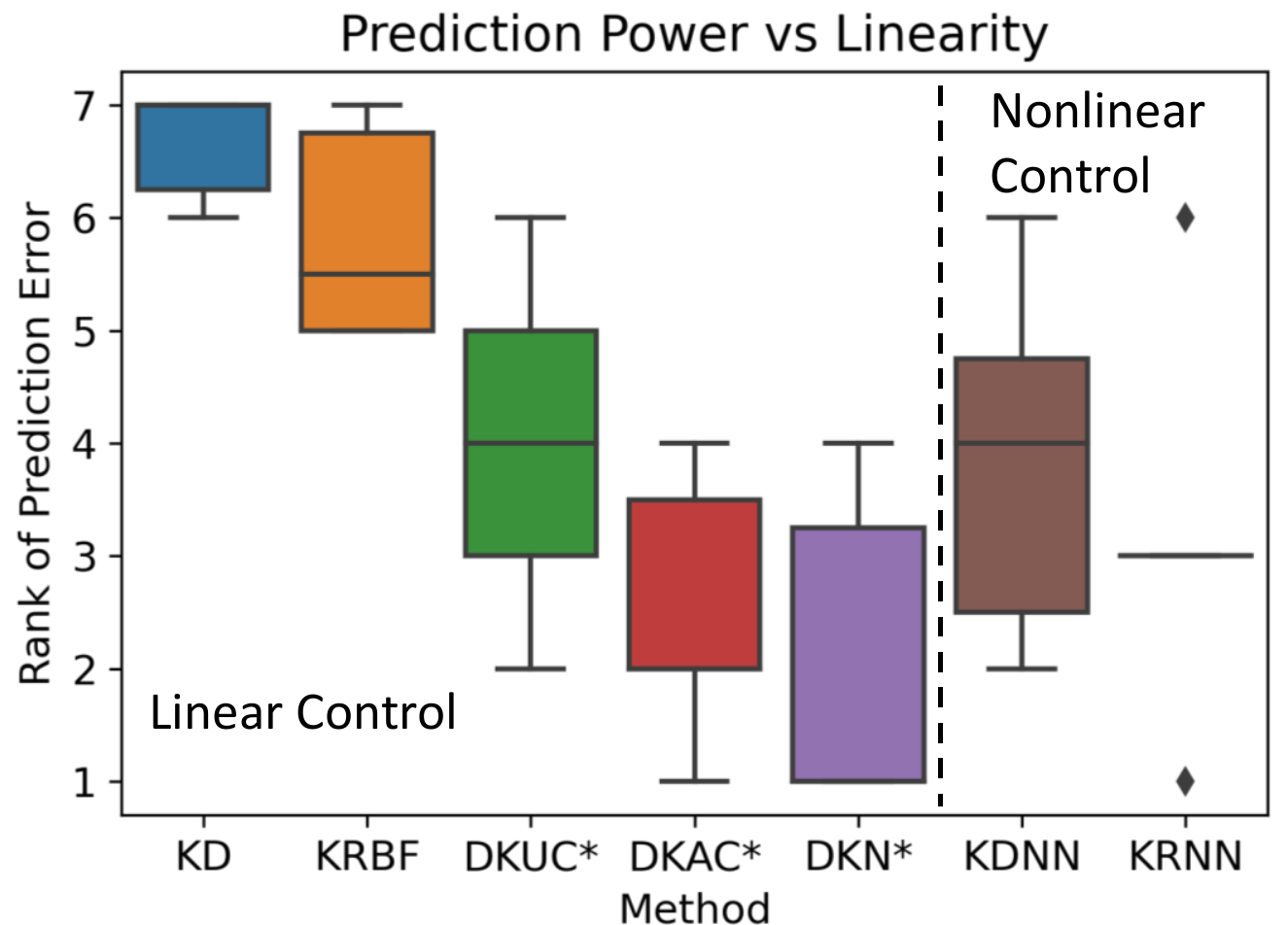}
	\caption{	\revise{Prediction Power vs. Linearity. The rank of prediction error is computed by sorting the prediction error tested in six environments in ascending order, and lower rank represents smaller prediction error.  * means our methods. KD, KRBF, KDNN and KRNN represents Koopman Derivative\cite{mamakoukas2021derivative}, Koopman RBF\cite{korda2018linear}, Deep neural network and Deep recurrent network methods.}
	}
	\label{fig:PredictionvsControl}
	\vspace{-.5cm}
\end{figure}

Recently, the Koopman operator \revise{has shown} great potential to alleviate such difficulties for nonlinear control under unknown dynamics. As a data-driven approach, the Koopman operator can \revise{automatically} learn the system model with the least square method. Furthermore, the Koopman operator \revise{can map} nonlinear dynamics to linear systems in the embedding space. And the embedded linear system is ready to be controlled via linear methods like LQR\cite{bemporad2002explicit,mamakoukas2021derivative} and MPC\cite{camacho2013model,korda2018linear}. \revise{Still, selecting the embedding function to maintain prediction quality remains a tough task.} Recent approaches focus on learning the embedding functions with deep neural networks \cite{lusch2018deep,azencot2020forecasting} and then apply linear control methods\cite{li2019learning,han2020deep}. \revise{However,} after lifting the state to embedded space, the linear-quadratic form of the cost function can be distorted in the hidden space, increasing the subsequent control difficulty. Moreover, previous Koopman-based approaches only model nonlinear systems with linear control input, destroying the prediction quality for general nonlinear systems.

In this work, we propose an end-to-end deep learning framework to learn the Koopman embedding function and Koopman Operator together to alleviate such difficulties. \revise{Our proposed embedding function concatenates the original state and the neural network encodings so that the original state can be preserved for further control. Hence, the cost function in the embedding space is consistent with that in the original space.} Furthermore, we propose an auxiliary control network that both \revise{encodes} the state and control to model the nonlinear state-dependent term in control. Then the encoded term is considered the new control variable for linear control. After deriving the optimal control policy, the actual control input can be recovered from the encoded term. \revise{In this way, we maintain both the prediction power and the linear dynamics of the modeled system, suitable for linear control methods.}

To further evaluate our results, we conduct experiments on several nonlinear systems. For prediction, experimental results demonstrate that our approach outperforms other Koopman-based approaches and achieves comparable or even better performance than the deep recurrent neural network. Consequently, better prediction performance ensures our approach to achieve better control results in nonlinear dynamic environments like damping pendulum, CartPole, and \revise{the seven DOF} robotic manipulator.		\footnote[1]{Source Code: \href{https://github.com/HaojieSHI98/DeepKoopmanWithControl}{https://github.com/HaojieSHI98/DeepKoopmanWithControl}}

In summary, the main contributions of these papers are:
\begin{enumerate}
	\item We propose an end-to-end deep learning framework that trains the embedding function and Koopman Operator together, maintaining both the prediction quality and the consistency of cost function in embedded space.
	\item \revise{We design an auxiliary control network that models the state-dependent nonlinearity in control input, improving the prediction performance while retaining the linear dynamics of the modeled system so that it is feasible for linear control methods.}
	\item We evaluate our approach in several nonlinear dynamic systems and achieve better prediction and control performance than existing Koopman-based methods and deep learning-based approaches.
\end{enumerate}

\section{RELATED WORK}	

\textbf{Prediction of nonlinear systems:} For unknown dynamics, data-driven methods have been considered a leading tool \revise{for learning} the hidden nonlinear dynamics model. Dynamic model decomposition algorithms (DMD,EDMD) \cite{schmid2010dynamic,tu2013dynamic,williams2015data} seeks an optimal linear operator matrix to fit the evolution of nonlinear systems with spectral decomposition. Sparse identification of nonlinear dynamics algorithms \cite{brunton2016sparse,brunton2016discovering} aim to discover the governing equations from data of nonlinear dynamics via sparse regression of the dynamics matrix. Deep neural networks (DNNs) also play an important role in directly learning the hidden nonlinear dynamics from data\cite{kumar2019comparative}, especially the deep recurrent neural networks that take advantage of the temporal information of data\cite{gencay1997nonlinear}. \revise{And latent state space methods that learn state representation with deep generative models succeed in modeling and controlling nonlinear systems from raw pixels.\cite{watter2015embed,lange2010deep}} Koopman-based approaches first select \revise{the} embedding function to lift the original state into embedding space and then learn the linear operator with the least square method to fit the linear dynamics in the lifted space. \revise{To choose the embedding functions, standard methods utilize the radial basis function (RBF) networks as the representation function\cite{korda2018linear}. The selection of RBF centers significantly influences the approximation performance, and it requires tremendous manual effort.\cite{yousef2004locating}.}
Derivative-based Koopman\cite{mamakoukas2021derivative} uses higher-order derivatives of nonlinear dynamics as lift function, which achieves promising prediction performance, but it requires the knowledge of the derivatives of unknown dynamics. In contrast, deep learning approaches promote the learning of embedding functions with deep neural networks\cite{korda2018linear,azencot2020forecasting}, improving the prediction quality for a long time horizon. 

\textbf{Direct Nonlinear Control:} \revise{Nonlinear control methods can be sorted as direct control approaches and linearization approaches.} As for direct control approaches, they directly control the nonlinear model with backstepping feedback control laws\cite{fossen1998nonlinear} or optimal control such as iLQR\cite{li2004iterative} and NMPC\cite{allgower2012nonlinear}. Backstepping methods require the analytical form of dynamic equations, which is inaccessible for complex nonlinear systems. \revise{Moreover, direct optimal control methods demand tremendous computational time for high-dimensional systems, which is computationally prohibitive for real-time control.} Besides, RL-based methods train the neural network to approximate the optimal policy via policy gradient or value iteration\cite{haarnoja2018soft,schulman2017proximal,levine2013guided}. They can compute optimal policy in real-time with feedforward networks but still suffer from sample efficiency and generalization issues.

\textbf{Nonlinear Control Linearization:} Local linearization serves as the basic approach for nonlinear control. It first linearizes the nonlinear function at the current state and then controls the system with LQR or MPC\cite{sastry2013nonlinear}, but the stability is only guaranteed in a local region where the linearization approximates the nonlinear dynamics well. Koopman-based approaches\cite{korda2018linear,li2019learning,mamakoukas2021derivative,han2020deep} highly enlarge the stable area by globally mapping the nonlinear dynamics to linear dynamics in embedding space. After lifting the state space to embedding space, liner control methods like LQR and MPC are ready to be deployed for real-time control.

\section{BACKGROUND}
\subsection{Basics of Koopman Operator}
Given the nonlinear systems with discrete dynamics:
\begin{equation}
	s_{k+1} = F(s_k)
\end{equation}
Koopman Operator $\mathcal{K}$ is an infinite linear operator that evolves the embedding functions $g$ of the state:
\begin{equation}
	\mathcal{K}g(s) = g\circ F(s)
\end{equation}
Considering the control input with the assumption that $u_{k+1}=u_k$, \revise{the evolution flow follows} as:
\begin{equation}
	g(x_{k+1},u_{k+1}) = \mathcal{K} g(x_k,u_k)
	\label{eq:Koopman}
\end{equation}
where $x\in \mathbb{R}^n, u\in\mathbb{R}^m,$ and $g:\mathbb{R}^{n+m}\rightarrow\mathbb{R}^d$. Therefore, with the embedding function $g$ lifting the state space to embedding space, Koopman Operator $\mathcal{K}$ maps the nonlinear dynamics to linear dynamics. Readers can refer to \cite{brunton2021modern} for more details \revise{on} Koopman Operator theory.

Previous methods simplify \eqref{eq:Koopman} by linearizing the control term. They first seperate the embedding function into two parts that $g(x,u) = [g_x(x,u);g_u(x,u)]$, and assume that $g_x(x,u)=g_x(x), g_u(x,u)=u$. Then the Eq.\eqref{eq:Koopman} is simplified as:
\begin{equation}       
	\left[                 
	\begin{array}{c}  
		g_x(x_{k+1})\\  
		u_{k+1} \\ 
	\end{array}\right]
	=\left[
	\begin{array}{cc}  
		K_{xx} & K_{xu}\\  
		K_{ux} & K_{uu} \\ 
	\end{array}
	\right]
	\left[
	\begin{array}{c}  
		g_x(x_{k})\\  
		u_{k} \\ 
	\end{array}				
	\right ]                
\end{equation}
which can further be written as:
\begin{equation}
	g_x(x_{k+1}) = K_{xx}g_x(x_k)+K_{xu}u_k
	\label{eq:KoopmanU}
\end{equation}
\revise{However,}  such simplification inevitably damages the prediction performance of the nonlinear system and can further affect the control results.
\subsection{Learning Koopman Operator}
\revise{Data-driven methods are} deployed to learn the Koopman Operator $\mathcal{K}$. Given the dataset  $g_k=g(x_k,u_k), k=[1,2,...,N]$, the Koopman Operator $K_d$ is computed by optimizing the mean square loss:
\begin{equation}
	K_d = \argmin_{K_d}{\Sigma_{i=1}^{N-1}||g_{i+1}-K_dg_i||}
	\label{eq:LinearRegression}
\end{equation}
The least-square problem has a closed-form solution:
\begin{equation}
	K_d^* =  PG^\dag
\end{equation} 
where $P=\frac{1}{N-1}\Sigma_{i=1}^{N-1}g_{i+1}g_i^T$, $G=\frac{1}{N-1}\Sigma_{i=1}^{N-1}g_{i}g_i^T$, and $\dag$ is the Moore–Penrose pseudoinverse.
\begin{figure*}[ht]
	\vspace{-.3cm}
	\centering
	\includegraphics[width=\textwidth]{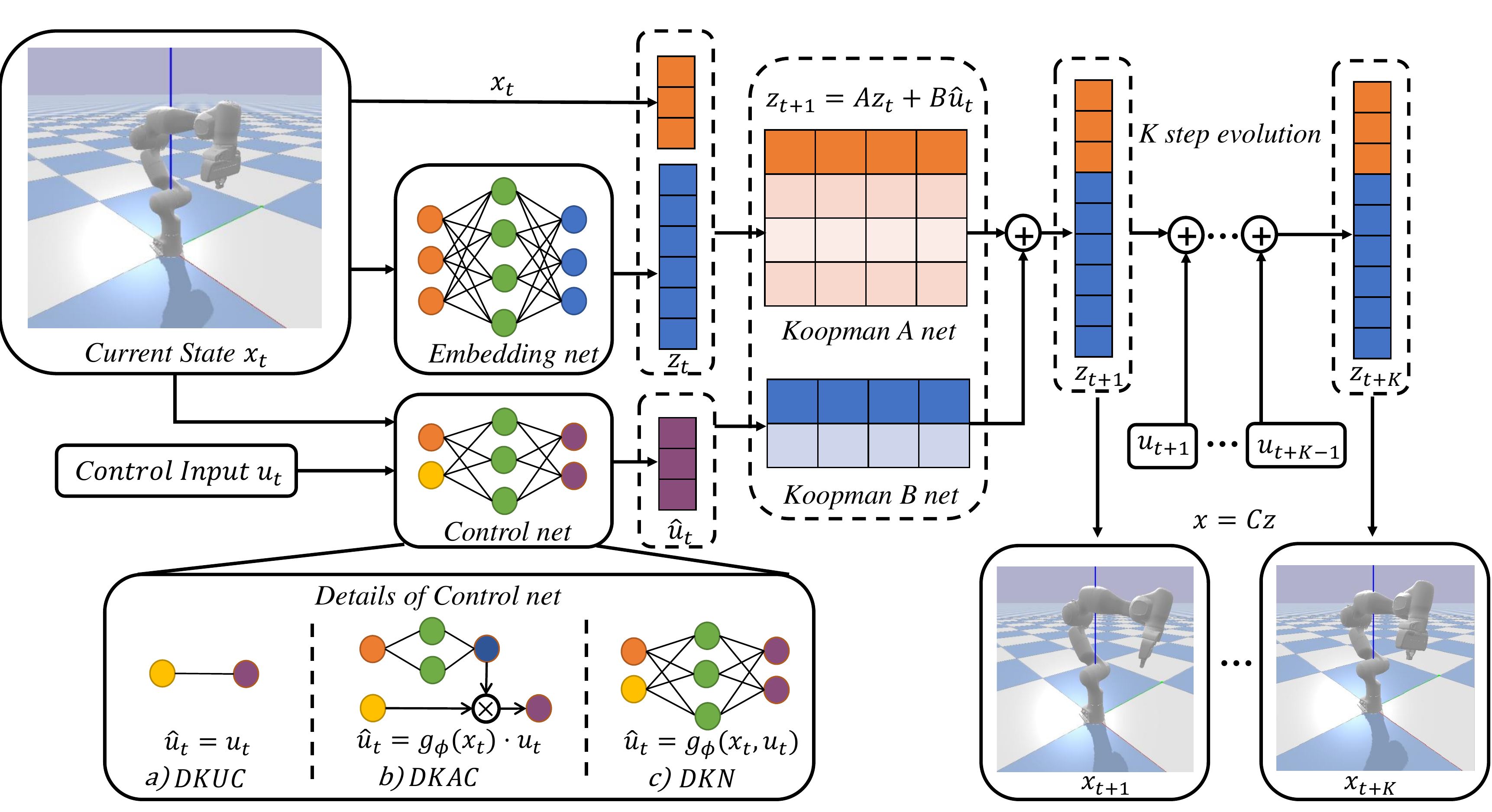}
	\caption{The overview of \revise{the} neural network framework for K-steps predictions.}
	\label{fig:prediction}
	\vspace{-.4cm}
\end{figure*}		
\section{METHODOLOGY}
\subsection{Framework}
We first seperate the embedding function into two parts: $g(x,u)=[g_x(x,u);g_u(x,u))]$. Without loss of generality, we assume that the first term is only related to the system state, so we get $g_x(x,u)=g_x(x)$. Based on Eq.\eqref{eq:Koopman}, it is derived that:
\begin{equation}
	g_x(x_{k+1}) = K_{xx}g_x(x_k)+K_{xu}g_u(x_k,u_k)
	\label{eq:KoopmanNonlinear}
\end{equation}

For the first part of embedding function, we use a deep neural network $\theta$  to parameterize $g_x(x,k)$. Instead of directly encoding the state like\cite{lusch2018deep}, which requires an extra decode network, we concatenate the origin state and network encoding together:
\begin{equation}
	z_k=g_x(x_k)=\left[
	\begin{array}{c}  
		x_{k}\\  
		g_{\theta}(x_k) \\ 
	\end{array}					 
	\right]
\end{equation}
where $z_k$ is the embedded state, $g_\theta:\mathbb{R}^n\rightarrow\mathbb{R}^d$ is the parameterized neural network. The advantage of this design is that the original state can be easily recovered by:
\begin{equation}
	x_k = Cz_k
	\label{eq:C}
\end{equation}
where the matrix $C\in\mathbb{R}^{n\times(n+d)}$ holds the simple form as:
\begin{equation}
	C=\left[ 
	\begin{array}{cc}
		I_n & 0\\
	\end{array}
	\right] 
\end{equation}
In this way, we can retain the form of the cost function for further control.

For the second part of embedding function, we have three versions to represent $g_u(x_k,u_k)$. The first version is called Deep KoopmanU with Control (DKUC) algorithm. It just simplifies the function as $g_u(x_k,u_k)=u_k$, then the evolution equation degrades to Eq.\eqref{eq:KoopmanU}. 

The sceond version is called Deep Koopman Affine with control (DKAC) algorithm, it considers the evolution function as control affine form, so that $g_u(x_k,u_k)=g_u(x_k)u$, and the Eq.\eqref{eq:Koopman} is modified as:
\begin{equation}
	z_{k+1} = K_{xx}z_k+K_{xu}g_u(x_k)u
\end{equation}
And we parameterize the function $g_u(x_k)$ by designing an \revise{auxilliary} control network $\phi$, so that $g_u(x_k)=g_\phi(x_k)$.

For the final version, we use the control network $\phi$ to approximate the function $g_u(x_k,u_k)$ directly as $g_u(x_k,u_k)=g_\phi(x_k,u_k)$, and the evolution function remains the same form as Eq.\eqref{eq:KoopmanNonlinear}. It is called Deep Koopman Nonlinear (DKN) algorithm.

\subsection{Feedforward Prediction}
Given the current state $x_t$, we can easily predict future K-steps states by the feedforward network. Besides the embedding function, we also parameterize the Koopman Operator matrix $K_xx$ and $K_xu$ by one layer linear network that $K_{xx}=A, K_{xu}=B$. Then the K-steps prediction follows:
\begin{gather}
	z_{t+k+1} = Az_{t+k}+Bg_\phi(x_{t+k},u_{t+k}), k=0,...,K-1 \notag \\
	z_t =g_x(x_t)= \left[
	\begin{array}{c}  
		x_{t}\\  
		g_{\theta}(x_t) \\ 
	\end{array}					 
	\right] \notag \\
	x_{t+k} = Cz_{t+k}
	\label{eq:forward}
\end{gather}
where $g_\phi(x,u)=u$ for KPUC algorithm, $g_\phi(x,u)=g_\phi(x)u$ for KPAC algorithm, and $g_\phi(x,u)=g_\phi(x,u)$ for KPN algorithm.
\subsection{K-steps Loss Function}
Standard Koopman-based algorithm first selects the embedding function and then learns the Koopman Operator matrix with linear regression for only step prediction. In this work, we learn the embedding function and Koopman Operator end-to-end instead, and we design \revise{the} K-steps prediction loss function for long-term prediction. Given the dataset $\left[X_i\in\mathbb{R}^{N\times n},U_i\in\mathbb{R}^{N\times m},i=0,1,2,...,K\right]$, we can compute the real embedded state $Z_i=g_x(X)$, and predict the K-steps sates $[\hat{Z}_i,i=1,2,...,K]$ in embedding space from the initial state $X_0$ via Eq.\eqref{eq:forward}. The loss function is \revise{computed as}:\revise{
	\begin{equation}
		L(\theta,\phi) = \Sigma_{i=1}^{K}\gamma ^{i-1}MSE(Z_i,\hat{Z}_i) 
		\label{eq:KstepLoss}
\end{equation}}where $\gamma$ is the weight decay hyper-parameter, and $MSE$ is the mean square loss function. Instead of only considering the following one-step prediction error, the K-steps loss focuses on the weighted sum of K-steps prediction error, conducive to prediction in a long time horizon.
\begin{figure}[h]
	\vspace{-.3cm}
	\centering
	\includegraphics[width=0.5\textwidth]{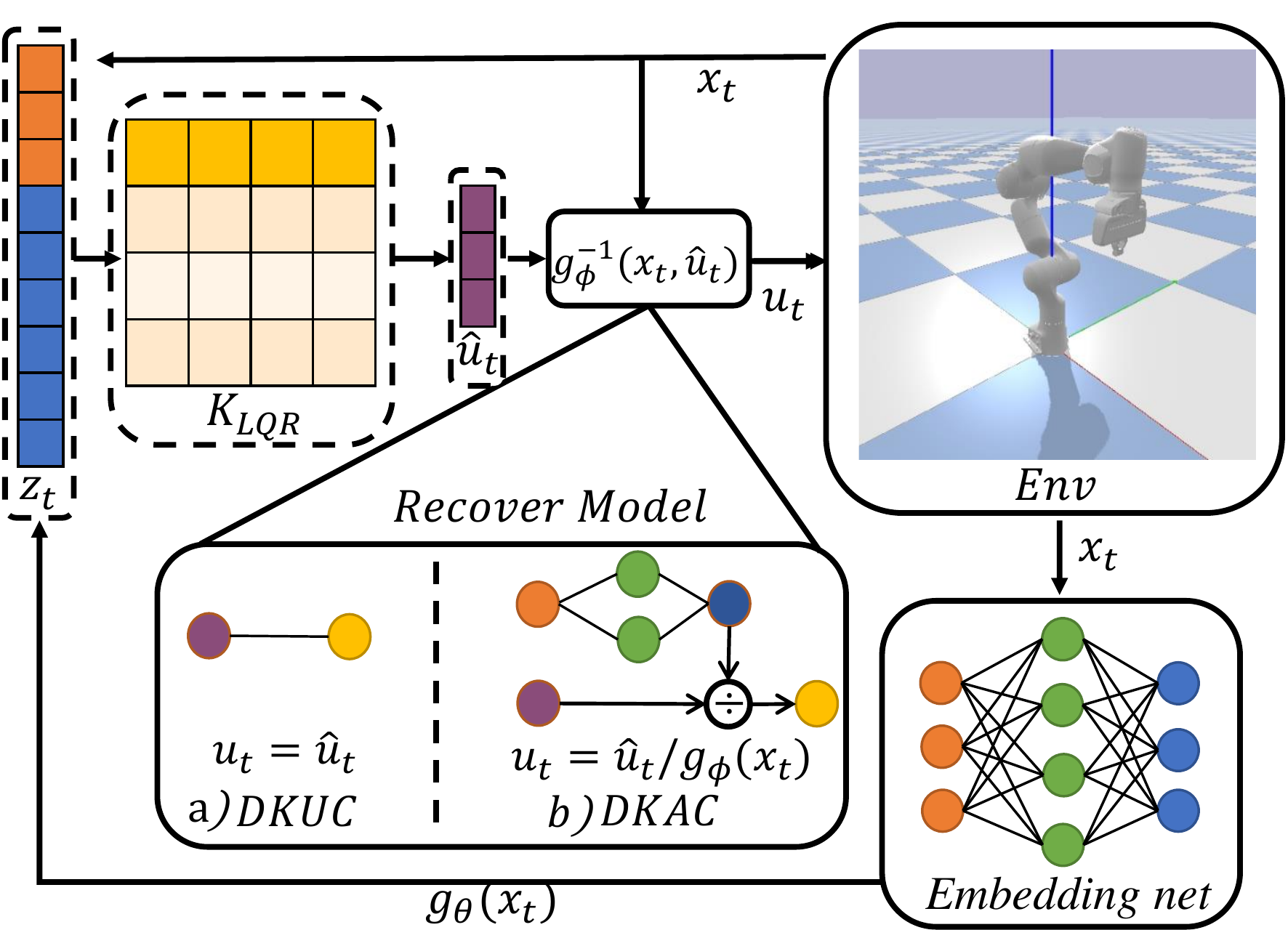}
	\caption{\revise{The} Framework of LQR Control with Deep Koopman Operator.}
	\vspace{-.4cm}
\end{figure}
\subsection{LQR Control}
Without loss of generality, we consider quadratic loss function and optimal control with infinite horizons and ignore constraints in this work. For general nonlinear systems, given the desired state $x_t^{des}$, the standard optimal control problem takes the form that:
\begin{gather}
	\min_{u_{t=1,...,\infty}}{\Sigma_{t=1}^{\infty}{(x_t-x_t^{des})}^TQ(x_t-x_t^{des})+u_t^TRu_t} \notag \\
	\text{s.t.  } x_t = F(x_{t-1},u_{t-1})
	\label{eq:LQR}
\end{gather}where $Q, R$ is the cost matrix related to state and control input. Since the evolution function is nonlinear, the optimal control problem is tough to be solved directly. 

On the contrary, the Koopman Operator maps the nonlinear systems to linear dynamics in embedding space, ensuring the feasibility of linear control methods. Therefore, we apply optimal control in embedding space and consider $g_\phi(x,u)$ as the new control variable instead, in which case it degrades to the LQR control problem. Based on Eq.\eqref{eq:forward},\eqref{eq:C}, the LQR problem can be rewritten as:
\begin{alignat}{2}
	\min_{\hat{u}_{t=1,...,\infty}}\quad & \Sigma_{t=1}^{\infty}{(z_t-z_t^{des})}^T\hat{Q}(z_t-z_t^{des})+\hat{u}_t^{T}\hat{R}\hat{u}_t \notag \\
	\text{s.t.} \quad & \begin{aligned}[t]
		z_{t+1} &= Az_t+B\hat{u}_t,\quad z_0=g_x(x_0) \\
		\hat{u}_t &= g_\phi(Cz_t,u_t)
	\end{aligned}\label{eq:KoopmanLQR}
\end{alignat} 
where $z_t^{des}=g_\theta(x_t^{des})$, $\hat{Q}=C^TQC$, $\hat{R}=R$ for KPUC algorithm, and $\hat{R}_t=g_\phi(x_t)^{-T}Rg_\phi(x_t)^{-1}$ for KPAC algorithm. Since $R$ is the pre-defined matrix and can be modified for different algorithms, we just assume $\hat{R}=R$ for simplification.

Eq.\eqref{eq:KoopmanLQR} has the closed-form solution via the LQR algorithm. After computing the LQR gain matrix $K_{LQR}$, the optimal control takes the form of:
\begin{equation}
	\hat{u}_t^*=K_{LQR}(z_t-z_t^{des})
\end{equation}
And we can recover the actual optimal control by:
\begin{equation}
	u_t^* = g_{\phi}^{-1}(x_t,\hat{u}_t^*)
\end{equation}
where $u_t^* = \hat{u}_t^*$ for KPUC algorithm, and $u_t^* = g_\phi(x_t)^{-1}\hat{u}_t^*$ for KPAC algorithm. The detailed algorithm is illustrated in Algorithm~\ref{alg:DKC}.

\begin{algorithm}[h]
	\caption{Deep Koopman with Control}
	\label{alg:DKC}
	\begin{algorithmic}[1]
		\REQUIRE $g_\theta$, embedding network \\ $g_\phi$, auxlliary control network \\ $A,B,C$, Koopman Operator matrix and recover matrix\\$Q,R$, cost matrix \\ $x_0, x_t^{des}$, initial state and desired trajectory\\ $T$, control steps\\$F$, nonlinear evolution equation
		\STATE initialize LQR gain matrix: $K_{LQR} \leftarrow LQR(A,B,Q,R)$
		\STATE embed the desired state $z_t^{des}\leftarrow g_x(x_t^{des})$
		\STATE reset the $env$ with initial state $x_0$
		\FOR{each t in \{0...T-1\}}
		\STATE embed current state $z_t \leftarrow g_x(x_t)$
		\STATE compute optimal control $\hat{u}_t^*\leftarrow K_{LQR}(z_t-z_t^{des})$
		\STATE recover actual optimal control $u_t^*\leftarrow g_\phi^{-1}(x_t,\hat{u}_t^*)$
		\STATE simulate the system forward $x_{t+1} \leftarrow F(x_t,u_t^*)$
		\ENDFOR
	\end{algorithmic}
\end{algorithm}

\section{EXPERIMENTS}
In this section, we conduct complete experiments to answer the following questions:

(1). How is the prediction performance of our approaches compared with previous Koopman-based approaches and deep learning-based approaches? \revise{And how is the sample efficiency of our proposed methods compared with others?}

(2). How is the control performance of our approaches compared with previous Koopman-based methods with linear control?

(3). Is our proposed control net conducive to nonlinear system prediction and control? Specifically, do our DPAC and DKN algorithms outperform the DKUC algorithm in prediction and control?  

\subsection{Environments}
\begin{figure*}
	\vspace{-.3cm}
	\centering
	\includegraphics[width=\textwidth]{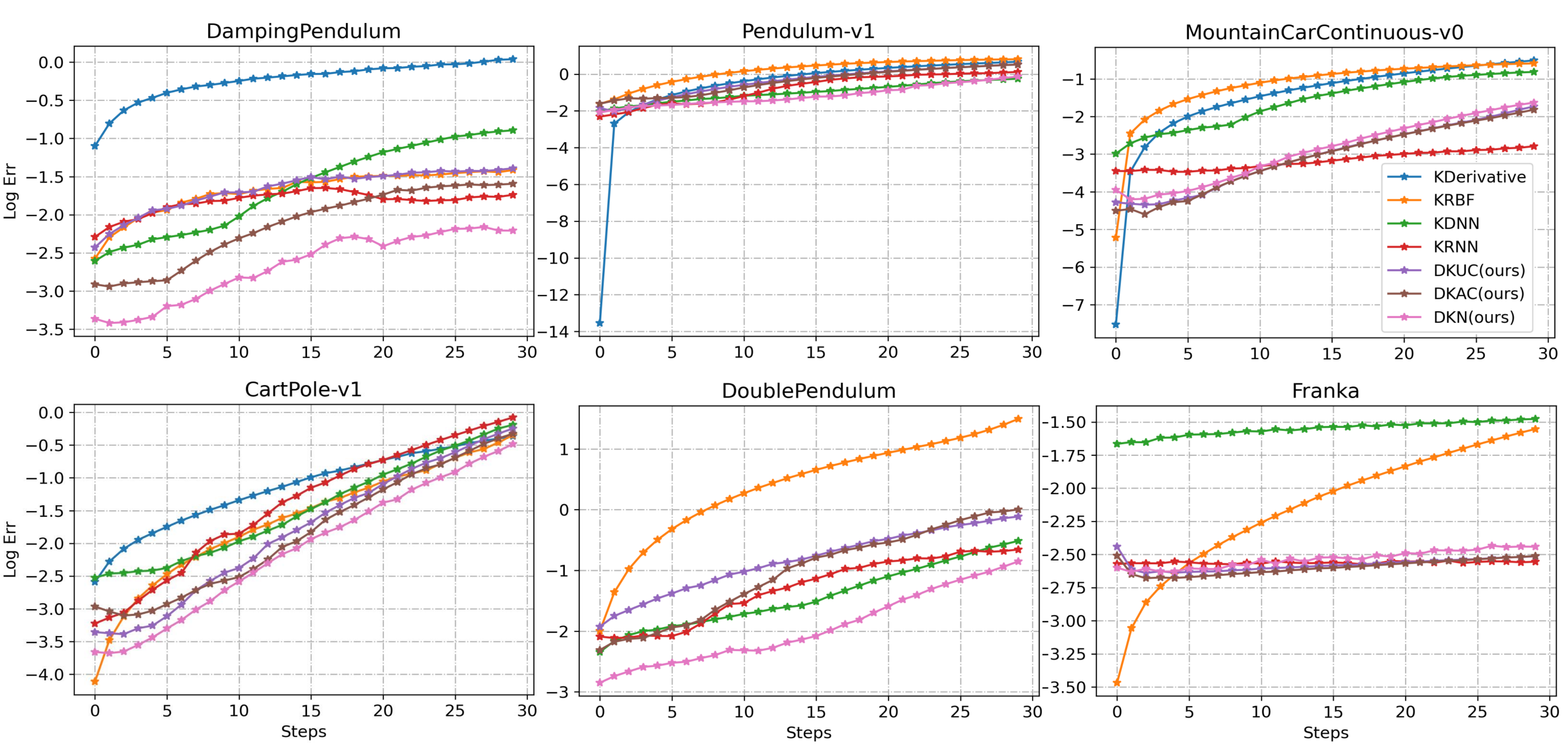}
	\caption{Prediction Results. \revise{The log of mean prediction error in the next thirty steps tested in six environments.}}
	\label{fig:PredictionResults}
\end{figure*}
To evaluate our approach, we conduct experiments on six nonlinear dynamics simulation envoriments, including (a) DampingPendulum, (b)  Pendulum, (c) MountainCarContious, (d) CartPole, (e) DoublePendulum, and (f) Franka. Among above six envrionments, b)-d) are environments  modified from OpenAI gym\cite{brockman2016openai}, and f) Franka is the same as \cite{mamakoukas2020memory}.

Besides, the dynamics of (a) DampingPendulum environment has the following form:
\begin{equation}  
	\frac{d}{dt}     
	\left[                 
	\begin{array}{c}  
		\theta\\  
		\dot{\theta} \\ 
	\end{array}\right]
	=\left[
	\begin{array}{c}  
		\dot{\theta}\\  
		-glsin\theta+\frac{1}{ml}b\dot{\theta}+\frac{1}{ml}cos\theta u\\
	\end{array}				
	\right ]                
\end{equation}
where $m$,$l$ is the mass and length of the pendulum, $b$ is the damping ratio, and $u$ is the external force put horizontally (the control input).

The dynamics of (e) holds the following partial derivative equation\cite{lynch2017modern}:
\begin{equation}
	M(\theta)\ddot{\theta}+c(\theta,\dot{\theta})+g(\theta)=u
\end{equation}
where $M(\theta)$ is the mass matrix, $c(\theta,\dot{\theta})$ is the velocity-product term, $g(\theta)$ is the gravity term and $u\in\mathbb{R}^2$ is the input force.
\subsection{K-steps Prediction}
We evaluate the prediction performance of our approaches on the above six environments and compare them with previous methods: Derivative-based Koopman (KDerivative)\cite{mamakoukas2021derivative}, RBF-based Koopman (KRBF)\cite{korda2018linear}, Deep neural networks (KDNN) and Deep recurrent networks (KRNN). For KDerivative and KRBF, we train them with linear regression as mentioned in Eq.\eqref{eq:LinearRegression}. For KDNN and KRNN, we generally train a neural network to approximate the nonlinear evolution function $F(x,u)$ with K-steps Loss in Eq.\eqref{eq:KstepLoss}. \revise{For DKAC and DKN, we choose the output dimension of their control net the same as the dimension of its original control input.} We use three hidden layers and 128 hidden units, the same learning rate, and 15-steps prediction loss for all deep learning-based approaches.

The final 30-steps prediction results is plotted in Fig. \ref{fig:PredictionResults}. We run four times for every environment and algorithm and plot the mean of log10 of the maximum error in 30-time horizons. And the errors at the 15th step are shown in Table \ref{tab:PredictionResults}. KDerivative requires the $N$th derivative of the dynamics, which is too complex or inaccessible for DoublePendulum and Franka, so we fail to conduct these two experiments for KDerivative. From the above figure and tables, we can find that:
(1). KDerivative and KRBF work better at the first several steps since the least square methods only optimize the first step prediction. On the contrary, deep learning-based approaches achieve better performance \revise{over} a long time horizon.
(2). \revise{The DKN} algorithm performs best and even better than KRNN in most cases, although KRNN takes advantage of the extra temporal information of data.
(3). Our algorithms DKUC and DKAC reduce the prediction error by magnitudes than previous Koopman-based approaches (KDerivative \& KRBF) and achieve comparable or even better results than fully deep learning-based approaches (KDNN \& KRNN).
(4). Our algorithm DKAC outperforms DKUC in prediction, especially on DampingPendulum, where the 15th step prediction error of DKAC is almost one-third of that of DKUC. It illustrates that our \revise{auxiliary} control net works for better prediction. 
\begin{table*}[]
	\vspace{-.3cm}
	\centering
	\caption{Prediction Results at the 15th Step}
	\label{tab:PredictionResults}
	\resizebox{\textwidth}{!}{%
		\begin{tabular}{ccccccc}
			\hline
			& DampingPendulum   & Pendulum          & MountainCar       & CartPole          & DoublePendulum    & Franka            \\ \hline
			KDerivative & 6.760e-1\textpm3.284e-2 		 & 9.634e-1\textpm1.936e-2 			& 6.725e-2\textpm3.524e-4 		  & 8.626e-2\textpm2.031e-3 &           -        &    -               \\
			KRBF        & 2.700e-2\textpm9.814e-4 		 & 2.624e+0\textpm9.173e-2 			& 1.233e-1\textpm9.268e-3		  & 2.854e-2\textpm2.031e-3 & 3.879e+0\textpm7.232e-2 & 8.639e-3\textpm8.657e-5 \\
			KDNN              & 2.517e-2\textpm4.462e-3 		 & 9.714e-2\textpm2.020e-2 			& 3.509e-2\textpm5.250e-3		  & 2.582e-2\textpm2.923e-3 & 2.636e-2\textpm1.486e-3 & 2.888e-2\textpm8.885e-4 \\
			KRNN              & 2.032e-2\textpm4.695e-3 	 	 & 2.990e-1\textpm8.550e-2 			& \textbf{6.027e-4\textpm2.128e-5} & 5.314e-2\textpm7.360e-3 & 6.392e-2\textpm9.584e-3 & 2.726e-3\textpm1.187e-4 \\
			DKUC(ours)        & 2.835e-2\textpm5.166e-3 		 & 5.982e-1\textpm3.138e-2 			& 9.714e-4\textpm7.404e-5 		  & 1.595e-2\textpm1.042e-3 & 1.510e-1\textpm1.005e-2 & 2.586e-3\textpm3.630e-4 \\
			DKAC(ours)        & 9.511e-3\textpm6.465e-4 		 & 5.153e-1\textpm1.809e-2 			& 9.663e-4\textpm4.724e-5 		  & 1.075e-2\textpm7.291e-4 & 1.304e-1\textpm2.537e-2 & \textbf{2.485e-3\textpm6.110e-4} \\
			DKN(ours)         & \textbf{2.575e-3\textpm8.424e-4} & \textbf{5.026e-2\textpm2.677e-3}  & 1.341e-3\textpm4.530e-4 		  & \textbf{8.423e-3\textpm8.117e-4} & \textbf{7.247e-3\textpm4.099e-3} & 2.995e-3\textpm1.705e-4 \\ \hline
		\end{tabular}%
	}
\end{table*}
\subsection{\revise{Sample Efficiency of Learning Algorithms}}
\begin{figure}[ht]
	\vspace{-.3cm}
	\centering
	\includegraphics[width=0.5\textwidth]{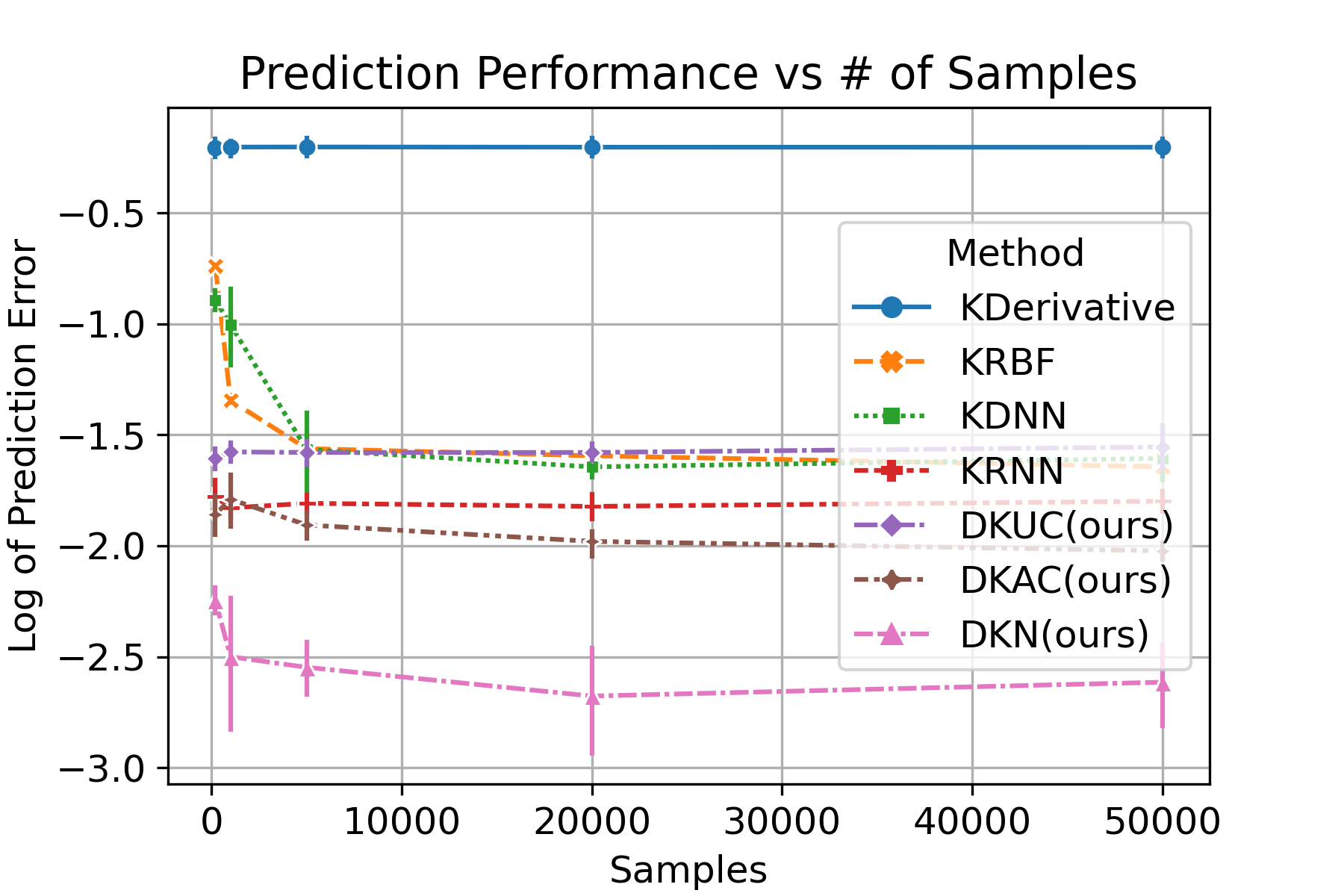}
	\caption{\revise{The mean and std of Log10  Error at the 15th step in the DampingPendulum environment with different numbers of training samples.}}
	\label{fig:Sample}
	\vspace{-.6cm}
\end{figure}

\revise{To evaluate the sample efficiency of our methods, we train them with various numbers of trajectory samples in the DampingPendulum environment, and the results are plotted in Fig \ref{fig:Sample}. The trajectory samples vary in [200,1000,5000,20000,50000], and each trajectory contains 15 time steps of data. From the figure, we can see that the performance of DKUC and DKAC algorithms is not sensitive to the number of training samples compared with KDNN and KRBF methods. Moreover, our proposed methods can achieve good prediction results after consuming 5000 trajectory samples (25 minutes of data).}  
\begin{figure}[ht]
	\vspace{-.5cm}
	\centering
	\includegraphics[width=0.48\textwidth]{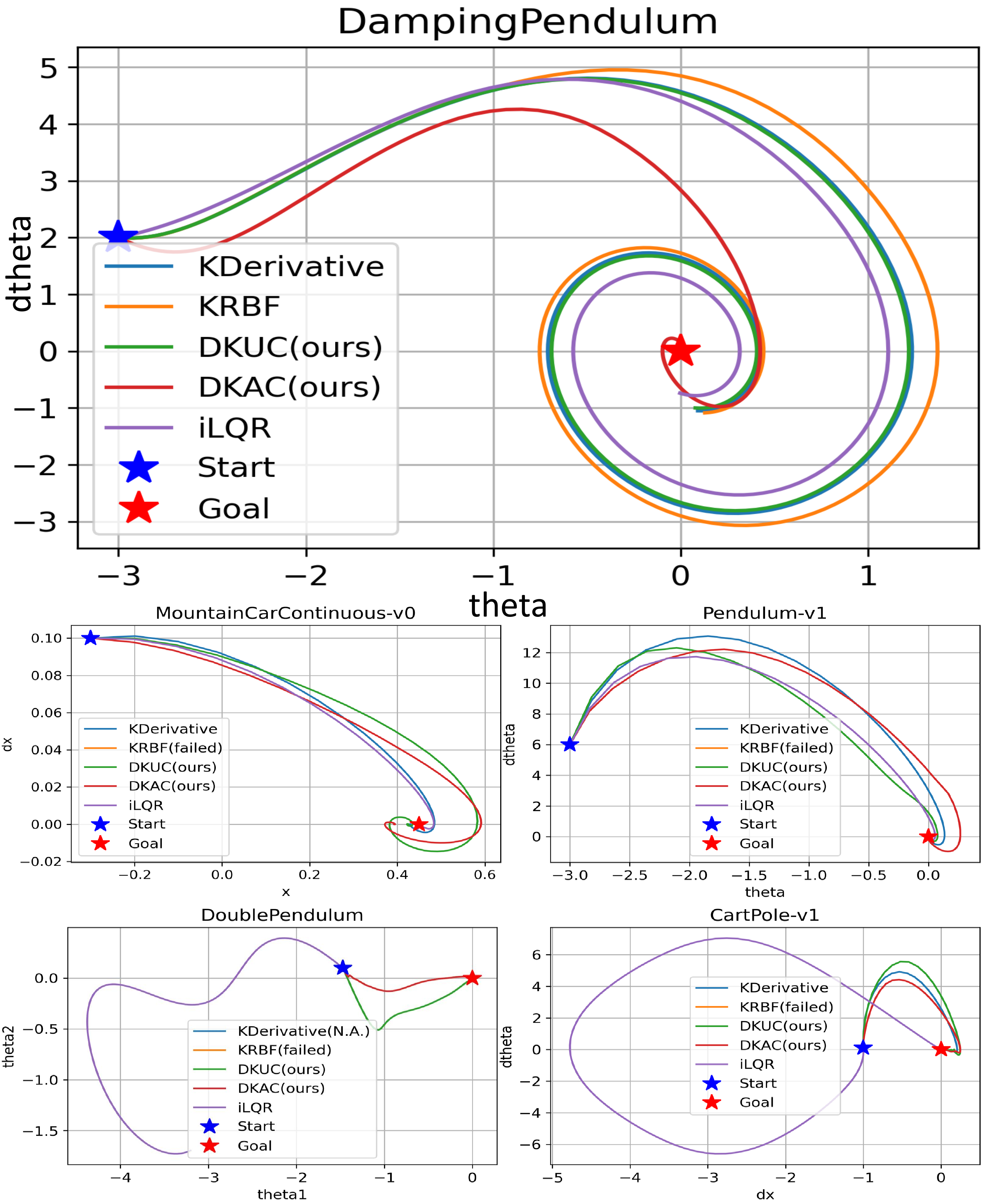}
	\caption{ \revise{Phase portraits of classical control results in six environments. The blue star mark means the initial state, and the red star mark represents the goal state. Control results of the KRBF algorithm running out of range are not plotted. KDerivative does not apply to DoublePendulum without access to the high order of derivatives.} }
	\label{fig:ControlResults}
	\vspace{-.3cm}
\end{figure}
\subsection{LQR Control}
As for control performance, we compare our approaches with KRBF and KDerivative, which are suitable for liner control methods like LQR, while KDNN and KRNN methods are fully nonlinear, and linear control methods are inapplicable to them. \revise{Moreover, we also added the iLQR\cite{li2004iterative} algorithm as a baseline for nonlinear control methods.} Besides, since we have \revise{not} found an invertible $g_\phi(x,u)$ of the DKN algorithm, we only deploy our proposed DKUC and DKAC algorithms for control.
\begin{figure}[ht]
	\centering
	\includegraphics[width=0.48\textwidth]{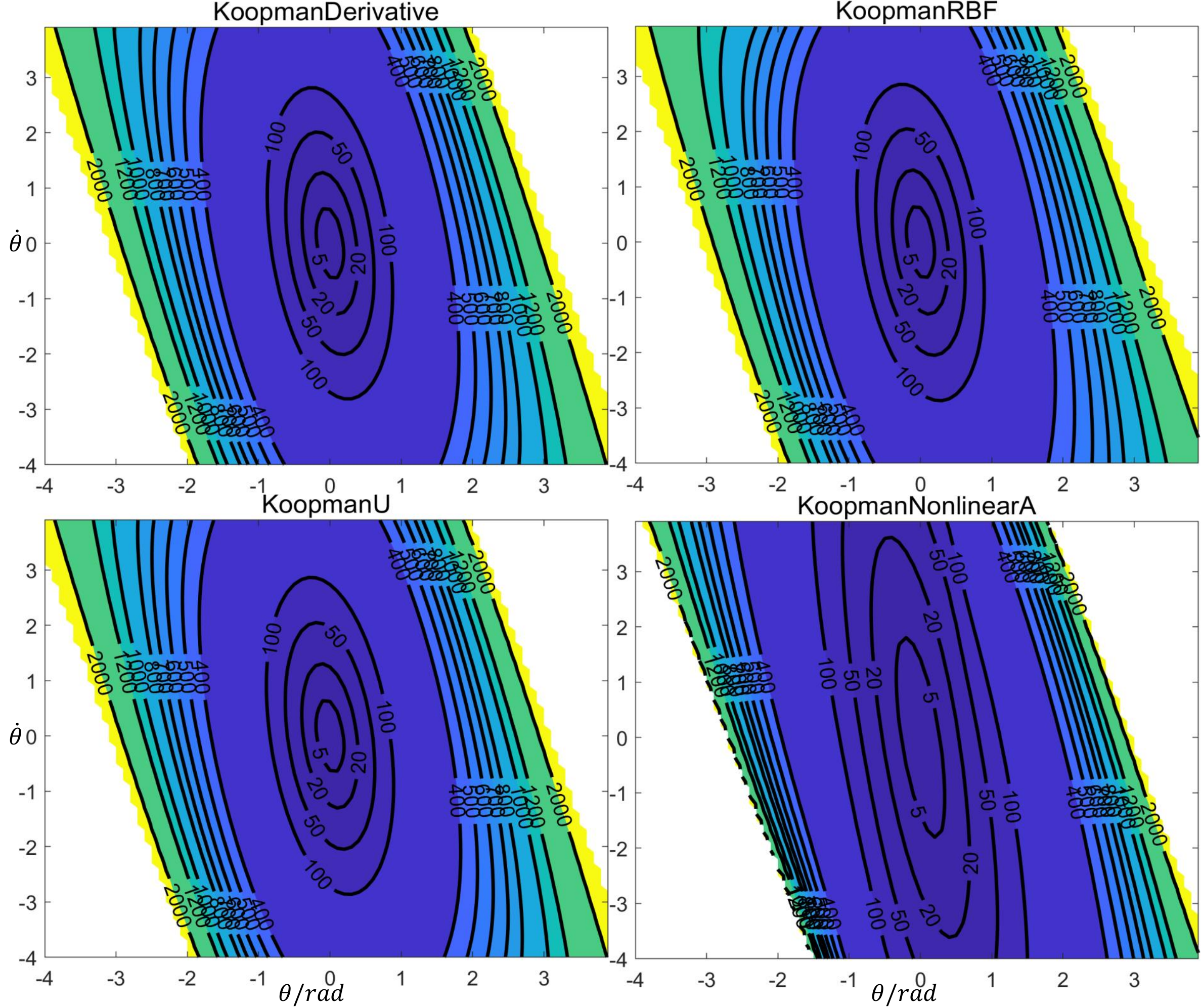}
	\caption{\revise{The total cost of all initial states in the DampingPendulum environment. Each point in the plane represents the initial state $(\theta,\dot{\theta})$, and the color represents the corresponding value.}}
	\label{fig:DampingPendulumCost}
\end{figure}
\begin{figure}[ht]
	\centering
	\includegraphics[width=0.5\textwidth]{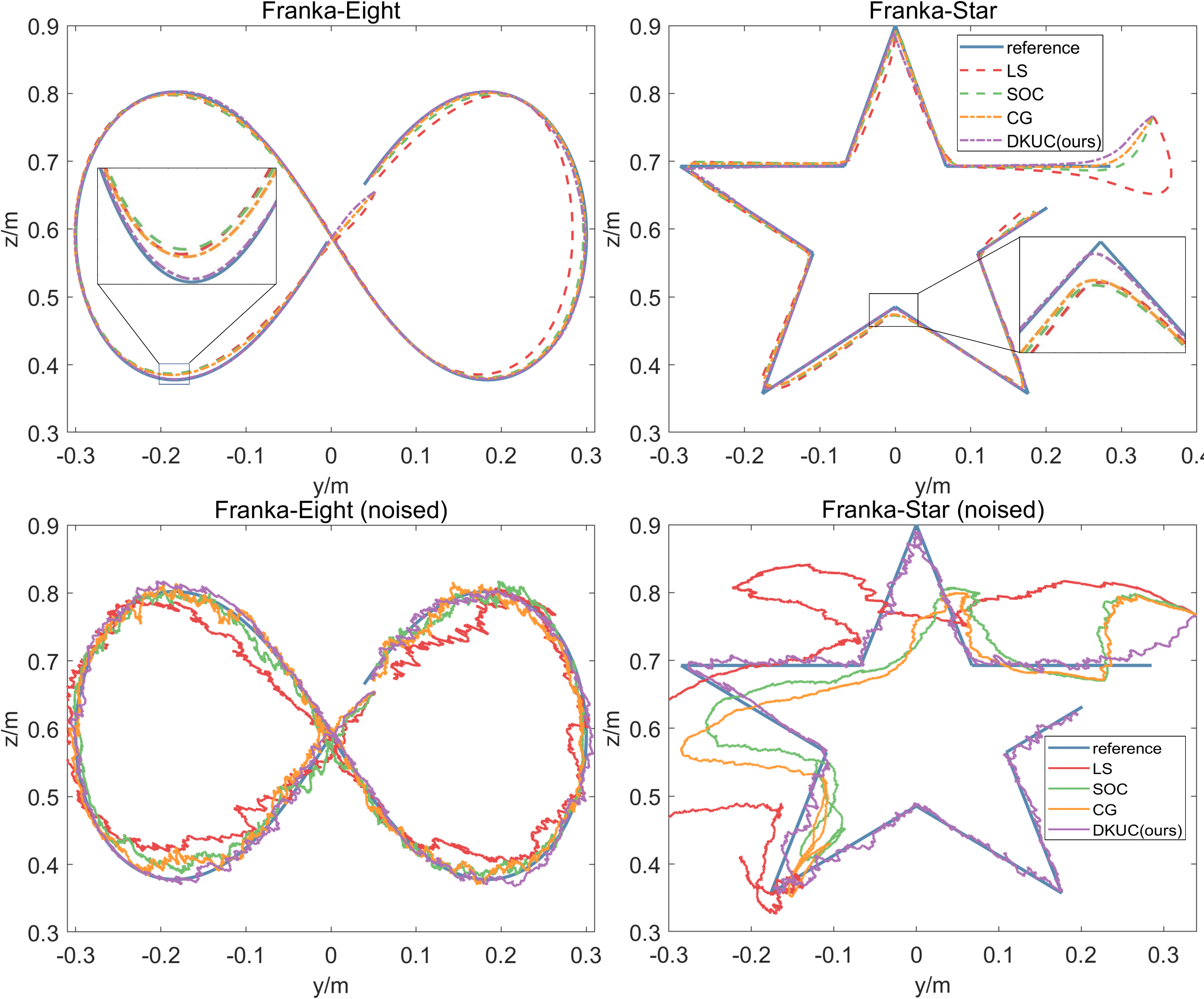}
	\caption{Franka Control Results. The desired trajectory of the robot end-effector generated in the y-z plane is plotted as the blue line. Trajectories of WLS and KRBF are out of range. Top Tracking results without measuring noise. Down Tracking results with Gaussian noise of $\mathcal{N}(0,0.01*I)$ added to the position of the robot end-effector.}
	\label{fig:FrankaControl}
	\vspace{-.3cm}
\end{figure}
\subsubsection{Classical Control}
\begin{table}[]
	\renewcommand\arraystretch{1.5}
	\centering
	\caption{The Total Costs of Classical Control}
	\resizebox{0.5\textwidth}{!}{%
		\begin{tabular}{@{}cccccc@{}}
			\hline
			& KDerivative & KRBF & DKUC             & DKAC &iLQR             \\ \hline
			DampingPendulum       & 1196.692          & 1265.189   & 1183.103         & \textbf{1053.759} & 1072.974 \\
			Pendulum              & 307.728           & NaN        & 305.387 & 309.232       & \textbf{301.101}    \\
			MountainCar			 & 15.933            & NaN        & 13.873  & 16.372       & \textbf{11.389}     \\
			CartPole              & 90.827            & NaN        & 100.781          & \textbf{89.878} & 5265.195  \\
			DoublePendulum        & -                 & NaN        & 2919.976         & \textbf{2905.536} & 11774.169 \\ \hline
		\end{tabular}%
	}
	\label{tab:Total Cost}
	\vspace{-3mm}
\end{table}
\begin{table}[ht]
	\renewcommand\arraystretch{1.1}
	\centering
	\caption{The Tracking Error in Franka Control}
	\resizebox{0.5\textwidth}{!}{%
		\begin{tabular}{@{}ccccc@{}}
			\hline
			& Franka-Eight         & Franka-Star & Franka-Eight(noised) & Franka-Eight(noised)         \\ \hline
			LS         & 2.968\textpm90.051          & 3.395\textpm90.289  & 7.505\textpm 0.339 & 93.217\textpm 0.771        \\
			SOC        & 2.588\textpm90.060          & 2.918\textpm90.124  & 4.611\textpm 0.192 & 73.006\textpm 5.109         \\
			WLS        & 233.834\textpm914.601       & 262.825\textpm916.344 & 237.889\textpm 16.652 & 461.962\textpm 44.285      \\
			CG         & 2.179\textpm90.661          & 2.496\textpm90.118  & 3.137\textpm 0.088 & 71.110\textpm 5.655         \\
			KRBF & 359.744\textpm9276.283      & 8165\textpm922782  & 352.152\textpm 156.918 & 3483\textpm 7038         \\
			DKUC       & \textbf{1.450\textpm90.074} & \textbf{1.775\textpm90.161} & \textbf{2.035\textpm 0.052} & \textbf{2.312\textpm 0.151} \\ \hline
		\end{tabular}%
	}
	\label{tab:Franka Err}
	\vspace{-4mm}
\end{table}	
Environments from (a) to (e) are classical control problems with force as input. \revise{The phase portrait  control results are plotted in Fig. \ref{fig:ControlResults}.} Table \ref{tab:Total Cost} shows the total cost defined in Eq. \eqref{eq:LQR}, where NaN means that the control cost is out-of-range.

\revise{From the above plot and table, it can be seen that our approaches, DKUC and DKAC, succeed in controlling all the systems and achieving good performance. KRBF is quite unstable since it is highly dependent on the selection of centers of RBF functions\cite{yousef2004locating}. Furthermore, we can find that DKAC outperforms DKUC and KDerivative in environments like DampingPendulum, where the control term is highly nonlinear and related to the state. Only DKAC succeeds in controlling the damping pendulum to achieve the zero point. iLQR method achieves the best performance in low dimensional tasks like Pendulum-v1, but it consumes more computing time and fails in higher dimensional tasks such as CartPole-v1 and DoublePendulum.}

To further illustrate the advantage of DKAC, we conduct control experiments \revise{in the DampingPendulum environment of} all the initial conditions $\left\lbrace (\theta,\dot{\theta})|\theta\in[-4,4],\dot{\theta}\in[-4,4]\right\rbrace $, and we plot the total cost \revise{given the corresponding initial states} in Fig.\ref{fig:DampingPendulumCost}. From the above plots, we can see that our DKAC algorithm succeeds in directing the system to the desired state with less cost in most initial conditions, while other methods fail when the initial state is far away from the desired state.

\subsubsection{Franka Control}
For Franka Environment, our control input is the desired \revise{joint} velocity \revise{at} 100Hz, and the actual force is computed via PID \revise{at} 1000Hz. Our control goal is to make the robot track the desired trajectory \revise{of the robot end-effector} $x_t^{des},t=1,2,...,N$ . The control term has \revise{seven DOF}, and it \revise{is} hard for us to design the cost matrix $R$ for DKAC. Since the derivative is not available for KDerivative, we only compare our DKUC algorithm with KoopmanRBF and methods in previous work\cite{mamakoukas2020memory}, including least square (LS), weighted least square (WLS), constraint generation (CG), and SOC. The Franka control results are plotted in Fig.\ref{fig:FrankaControl} and we compare the tracking error $\Sigma_{t=1}^{N}||x_t-x_t^{des}||_2$ in 10 random initial states in Table \ref{tab:Franka Err}. \revise{Furthermore, we add Gaussian noise $X\in \mathcal{N}(0,0.01*I)$ to the position of the robot end-effector to evaluate the robustness of the algorithms. Based on the experimental results, we can see that our algorithm achieves the lowest tracking error and is robust in measuring noise. Especially for Franka-Star (noised) task, only our algorithm succeeds in tracking the desired trajectory.}
\section{\revise{LIMITATION AND FUTURE DIRECTION}}
\revise{ Although our proposed algorithms outperform other state-of-the-art methods in prediction and control from the experimental results, this work still has several limitations. Firstly, though the DKN algorithm achieves the best prediction performance, it is hard to find a unique inverse mapping function to recover the real control signals from the output of its auxiliary control network. To address this difficulty, designing an invertible network to decode the output of the auxiliary control network would be the key problem. One possible way could be to apply the autoencoder \cite{tschannen2018recent} to decode the output. Training an invertible neural network \cite{behrmann2019invertible} directly for the auxiliary control network would also be a possible solution. Secondly, expanding our approaches to higher dimensional nonlinear systems is quite promising for future study, especially for soft robot systems whose dynamics are too complex to be derived. Last but not least, our approach does not consider the input and state constraints. Since our embedding state contains the original state, the state constraints are easy to derive in the embedding space. However, the input constraints are distorted by the control network of DKAC and DKN algorithms in the embedding space. Therefore, our future study will also focus on dealing with the input constraints in the embedding space.  }
\section{CONCLUSION}
In this work, we propose an end-to-end deep learning framework to learn the Koopman embedding function and Koopman Operator together, improving the prediction power of nonlinear systems and the control performance with LQR. Furthermore, we design a control network to model the nonlinear state-dependent term related to control, enhancing the prediction performance by strengthening the expressive power \revise{and} gaining even better prediction quality than the recurrent neural network. Experimental results prove that this approach is also conducive to the control of general systems with high nonlinearity on control terms. We hope that our work could facilitate the further study of real-time control on complex nonlinear systems.

{\small
	\bibliographystyle{plain}
	\bibliography{myref}
}
\vfill

\end{document}